\documentclass[11pt,letterpaper]{article}
\usepackage[letterpaper]{geometry}
\usepackage{acl2012}
\usepackage{times}
\usepackage{latexsym}
\usepackage{amsmath}
\usepackage{amssymb}
\usepackage{multirow}
\usepackage{graphicx}
\usepackage{algorithm}
\usepackage{booktabs}
\usepackage{color}
\usepackage{subcaption}
\usepackage{afterpage}
\usepackage{anyfontsize}
\usepackage[noend]{algpseudocode}
\usepackage{url}
\usepackage{enumitem}
\usepackage{breqn}
\makeatletter
\newcommand{\@BIBLABEL}{\@emptybiblabel}
\newcommand{\@emptybiblabel}[1]{}
\makeatother
\usepackage[hidelinks]{hyperref}
\usepackage{tabularx}
\usepackage[framemethod=tikz]{mdframed}
\usepackage{bm}
\DeclareMathOperator*{\argmax}{arg\,max}
\definecolor{boxcolor}{rgb}{0,0,0} 

\newmdenv[innerlinewidth=0.5pt, roundcorner=4pt,linecolor=boxcolor,innerleftmargin=6pt,
innerrightmargin=6pt,innertopmargin=6pt,innerbottommargin=6pt]{annotationbox}

\title{Representation Learning for Grounded Spatial Reasoning}

\author{Michael Janner, Karthik Narasimhan, and Regina Barzilay \\
Computer Science and Artificial Intelligence Laboratory \\
Massachusetts Institute of Technology \\
{\tt \{janner, karthikn, regina\}@csail.mit.edu}}

\date{}

\begin{document}
\maketitle

\begin{abstract}
The interpretation of spatial references is highly contextual, requiring joint inference over both language and the environment. We consider the task of spatial reasoning in a simulated environment, where an agent can act and receive rewards. The proposed model learns a representation of the world steered by instruction text.  This design allows for precise alignment of local neighborhoods with corresponding verbalizations, while also handling global references in the instructions. We train our model with reinforcement learning using a variant of generalized value iteration. The model outperforms state-of-the-art approaches on several metrics, yielding a 45\% reduction in goal localization error.
\footnote{Code and dataset are available at \url{https://github.com/JannerM/spatial-reasoning}}

\end{abstract}

\section{Introduction}
\label{sec:introduction}

\begin{figure}[!t]
  \centering 
  \includegraphics[width=1.0\linewidth]{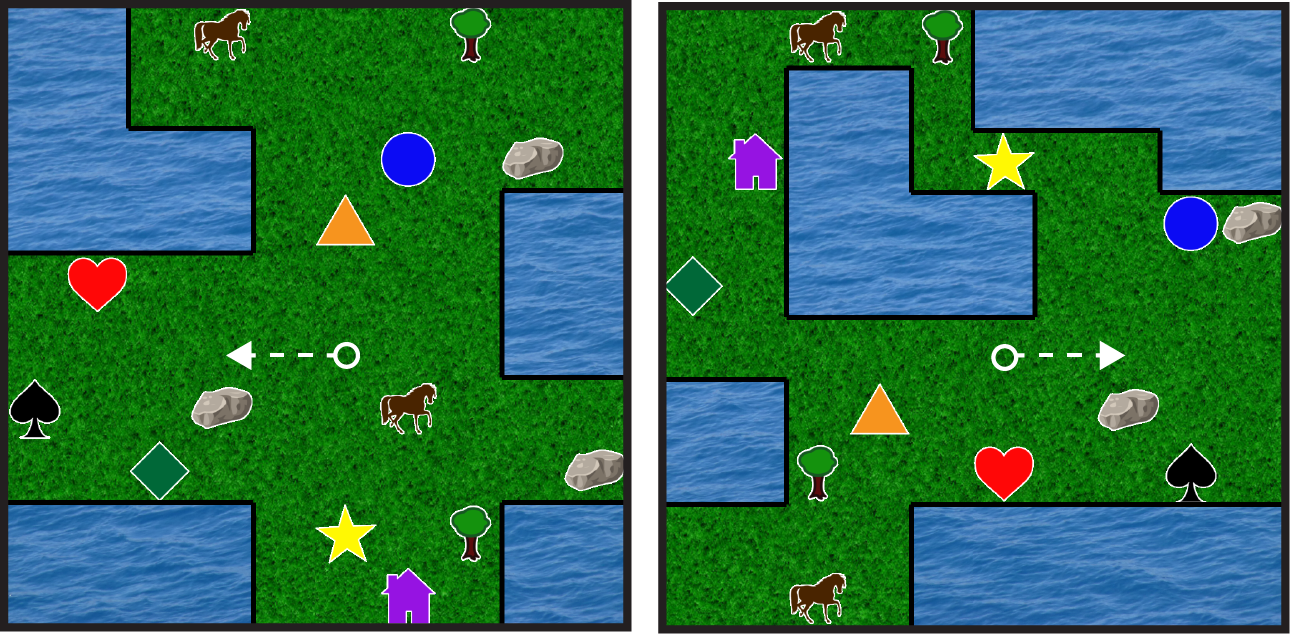}
  
  \begin{tabular}{ c }
  \\
  \emph{Reach the cell above the westernmost rock} \\  
\end{tabular}
  \caption{Sample 2D worlds and an instruction describing a goal location. The optimal path from a common start position, denoted by a white dashed line, varies considerably with changes in the map layout.}
	\label{fig:example}
\end{figure}

Understanding spatial references in natural language is essential for successful human-robot communication and autonomous navigation. This problem is challenging because interpretation of spatial references is highly context-dependent. For instance, the instruction \emph{``Reach the cell above the westernmost rock''} translates into  different goal locations in the two environments shown in Figure~\ref{fig:example}. Therefore, to enable generalization to new, unseen worlds, the model must jointly reason over the instruction text and environment configuration. Moreover, the richness and flexibility in verbalizing spatial references further complicates interpretation of such instructions. 

In this paper, we explore the problem of spatial reasoning in the context of interactive worlds. Specifically, we assume access to a simulated environment, in which an agent can take actions to interact with the world and is rewarded for reaching the location specified by the language instruction. This feedback is the only source of supervision the model uses for interpreting spatial references. 

The key modeling task here is to induce a representation that closely ties environment observations and linguistic expressions.  In prior work, this issue was addressed by learning representations for each modality and then combining them, for instance, with concatenation~\cite{misra2017mapping}.  While this approach captures high-level correspondences between instructions and maps, it does not encode detailed, lower-level mappings between specific positions on the map and their descriptions. As our experiments demonstrate, combining the language and environment representations in a spatially localized manner yields significant performance gains on the task.


To this end, our model uses the instruction text to drive the learning of the environment representation. We start by converting the instruction text into a real-valued vector using a recurrent neural network with LSTM cells~\cite{hochreiter1997long}. Using this vector as a kernel in a convolution operation, we obtain an instruction-conditioned representation of the state. This allows the model to reason about immediate local neighborhoods in references such as \emph{``two cells to the left of the triangle''}. We further augment this design to handle global references that involve information concerning the entire map (e.g. \emph{``the westernmost rock''}). This is achieved by predicting a global value map using an additional component of the instruction representation. The entire model is trained with reinforcement learning using the environmental reward signal as feedback.

We conducted our experiments using a 2D virtual world as shown in Figure~\ref{fig:example}. Overall, we created over 3,300 tasks across 200 maps, with instructions sourced from Mechanical Turk. We compare our model against two state-of-the-art systems adapted for our task~\cite{misra2017mapping,schaul2015universal}. The key findings of our experiments are threefold. First, our model can more precisely interpret instructions than baseline models and find the goal location, yielding a 45\%  reduction in Manhattan distance error over the closest competitor. Second, the model can robustly generalize across new, unseen map layouts. Finally, we demonstrate that factorizing the instruction representation enables the model to sustain high performance when handling both local and global references.

\section{Related Work}
\label{sec:related_work}

\paragraph{Spatial reasoning in text} This topic has attracted both theoretical and practical interest. From the linguistic and cognitive perspectives, research has focused on the wide range of mechanisms speakers use to express spatial relations~\cite{tenbrink2007space,viethen2008use,byrne1989spatial,li2002turning}. The practical implications of this research are related to autonomous navigation~\cite{moratz2006spatial,levit2007interpretation,tellex2011understanding} and human-robot interaction~\cite{skubic2004spatial}. 

Previous computational approaches include techniques such as proximity fields~\cite{kelleher2006proximity}, spatial templates~\cite{levit2007interpretation} and geometrically defined mappings~\cite{moratz2006spatial,kollar2010toward}. More recent work in robotics has integrated text containing position information with spatial models of the environment to obtain accurate maps for navigation~\cite{walter2013learning,hemachandra2014learning}.  
Most of these approaches typically assume access to detailed geometry or other forms of domain knowledge. In contrast to these knowledge-rich approaches, we are learning spatial reference via interaction with the environment, acquiring knowledge of the environment in the process.

\paragraph{Instruction following} Spatial reasoning is a common element in many papers on instruction following~\cite{macmahon2006walk,vogel2010learning,chen2011learning,artzi2013weakly,kim2013adapting,Andreas15Instructions}. As a source of supervision, these methods assume access to demonstrations, which specify the path corresponding with provided instructions. In our setup, the agent is only driven by the final rewards when the goal is achieved. This weaker source of supervision motivates development of new techniques not considered in prior work.


 
More recently, \newcite{misra2017mapping} proposed a neural architecture for jointly mapping instructions and visual observations (pixels) to actions in the environment. Their model separately induces text and environment representations, which are concatenated into a single vector that is used to output an action policy. While this representation captures coarse correspondences between the modalities, it doesn't encode mappings at the level of local neighborhoods, negatively impacting performance on our task. 

\paragraph{Universal value functions}
The idea of generalized value functions has been explored before in ~\newcite{schaul2015universal}. The technique, termed UVFA, presents a clever trick of factorizing the value function over states and goals using singular value decomposition (SVD) and then learning a regression model to predict the low-rank vectors. This results in quick and effective generalization to all goals in the same state space. However, their work stops short of exploring generalization over map layouts, which our model is designed to handle. Furthermore, our setup also involves specifying goals using natural language instructions, which is different from the coordinate-style specification used in that work.
 

\section{General Framework}
\begin{figure*}[t]
    \centering
    \includegraphics[width=0.8\linewidth]{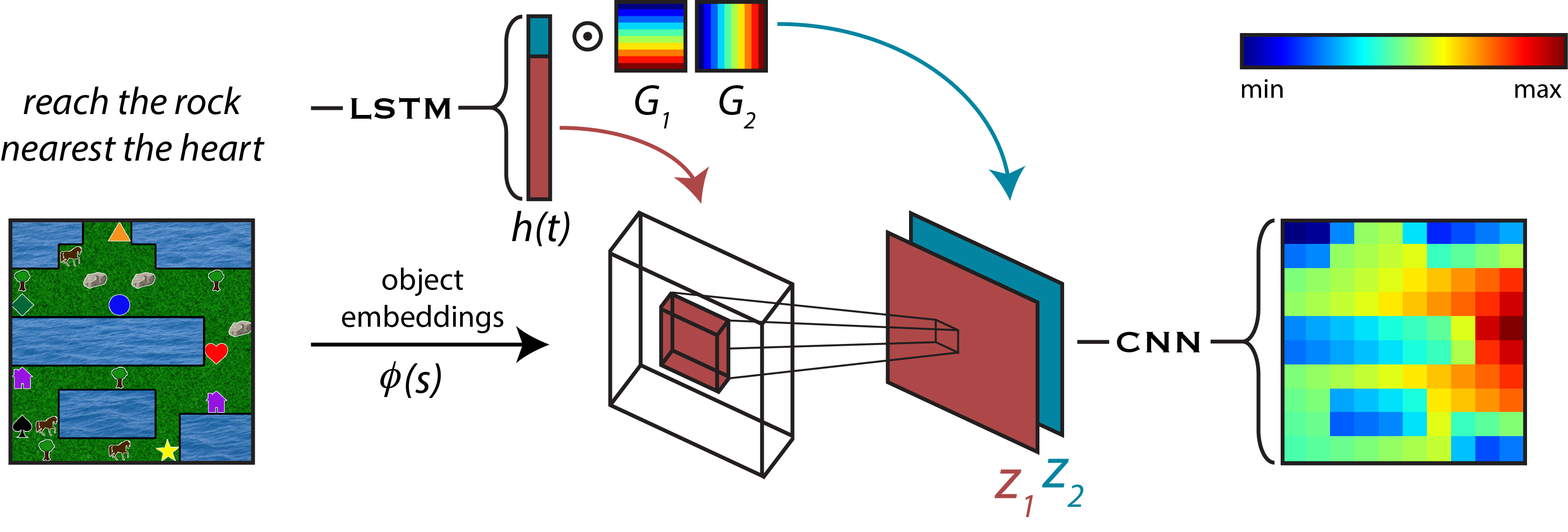}
    \caption{A schematic depiction of our model. Text instructions are represented as a vector $h(t)$ and states as embeddings $\phi(s)$. A portion of the text representation is used as a convolutional kernel on $\phi(s)$, giving a \emph{text-conditioned} local state representation $z_1$. The remaining components are used as coefficients in a linear combination of gradient functions to give a global map-level representation $z_2$. $z_1$ and $z_2$ are concatenated and input to a convolutional neural network to predict the final value map.}
    \label{fig:model}
\end{figure*}

\label{sec:task}
\paragraph{Task setup}
We model our task as a Markov Decision Process (MDP), where an autonomous agent is placed in an interactive environment with the capability to choose actions that can affect the world. A goal is described in text, and rewards are available to the agent correspondingly.
The MDP can be represented by the tuple $\langle S, A, X, T, R\rangle$, where $S$ is the set of all possible state configurations, $A$ is the set of actions available to the agent, $X$ is the set of all goal specifications\footnote{We will use the terms goal specifications and instructions interchangeably.} in natural language, $T(s' | s, a, x)$ is the transition distribution, and $R(s, x)$ is the reward function. A state $s \in S$ includes information such as the locations of different entities along with the agent's own position. 
In this work, $T$ is deterministic in the environments considered; however, our methods also apply in the stochastic case.  

\paragraph{Text instructions}
Prior work has investigated human usage of different types of referring expressions to describe spatial relations~\cite{levinson2003space,viethen2008use}. In order to build a robust instruction following system, we examine several categories of spatial expressions that exhibit the wide range of natural language goal descriptions. Specifically, we consider instructions that utilize objects/entities present in the environment to describe a goal location.  These instructions can be categorized into three groups: 
\begin{enumerate}[label=(\alph*)] 
\item Text referring to a specific entity. (e.g. \emph{Go to the circle.})
\item Text specifying a location using a single referent entity. (e.g. \emph{Reach the cell above the westernmost rock.})
\item Text specifying a location using multiple referent entities. (e.g. \emph{Move to the goal two squares to the left of the heart and top right of house.})
\end{enumerate}
These three categories exemplify an increasing level of complexity, with the last one having multiple levels of indirection.

In each category, we have both local and global references to objects. Local references require an understanding of spatial prepositional phrases such as `above', `in between' and `next to' in order to determine the precise goal location. This comprehension is invariant to the global position of the object landmark(s) provided in the instruction. A global reference, on the other hand, contains superlatives such as `easternmost' and `topmost', which require reasoning over the entire map. For example, in the case of (a) above, a local reference would describe a unique object\footnote{A local reference for a non-unique object would be ambiguous, of course.} (e.g. \emph{Go to the circle}), whereas a global reference might require comparing the positions of all objects of a specific type (e.g. \emph{Go to the northernmost tree}).

A point to note is that we do not assume any access to mapping from instructions to objects or entities in the worlds or a knowledge of spatial ontology  -- the system has to learn this entirely through feedback from the environment. 

\paragraph{Generalized Value Iteration}
Learning to reach the goal while maximizing cumulative reward can be done by using a \emph{value function} $V(s)$~\cite{sutton1998introduction} which represents the agent's notion of expected future reward from state $s$. A popular algorithm to learn an optimal value function is Value Iteration (VI)~\cite{bellman1957}, which uses the technique of dynamic programming. 

In the standard Bellman equation, the value function is dependent solely on state. \newcite{schaul2015universal} proposed a value function $V(s, g)$ describing the expected reward from being in state $s$ given goal $g$,
capturing that state values are goal-dependent and that a single environment can offer many such goals. We also make use of such a \emph{generalized value function}, although our goals are not observed directly as coordinate locations or states themselves but rather described in natural language.  With $x$ denoting a textual description of a goal, our VI update equations are:

\begin{dmath*}
Q(s, a, x) = R(s, x) + \gamma \sum_{s' \in S} T(s' | s, a, x) V(s', x) \\
\end{dmath*}
\begin{dmath}
V(s, x) = \max_a Q(s,a,x)
\end{dmath}
where $Q$ is the action-value function, tracking the value of choosing action $a$ in state $s$. Once an optimal value function is learned, a straightforward action policy is: 

\begin{dmath}
\label{eq:policy}
\pi(s,x) = \argmax_a Q(s,a,x)
\end{dmath}

\section{Model}

Generalization over both environment configurations and text instructions requires a model that meets two desiderata. First, it must have a flexible representation of goals, one which can encode both the local structure and global spatial attributes inherent to natural language instructions. Second, it must be compositional, in order to learn a generalizable representation of the language even though each unique instruction will only be observed with a single map during training. Namely, the learned representation for a given instruction should still be useful even if the objects on a map are rearranged or the layout is changed entirely.



To that end, our model combines the textual instructions with the map in a spatially localized manner, as opposed to prior work which joins goal representations and environment observations via simpler functions like an inner product \cite{schaul2015universal}. While our approach can more effectively learn local relations specified by language, it cannot naturally capture descriptions at the global environment level. To address this problem, we also use the language representation to predict coefficients for a basis set of gradient functions which can be combined to encode global spatial relations. 

More formally, inputs to our model (see Figure~\ref{fig:model}) consist of an environment observation $s$ and textual description of a goal $x$. For simplicity, we will assume $s$ to be a $2${\sc{D}} matrix, although the model can easily be extended to other input representations. We first convert $s$ to a $3${\sc{D}} tensor by projecting each cell to a low-dimensional embedding ($\phi$) as a function of the objects contained in that cell. In parallel, the text instruction $x$ is passed through an {\sc{LSTM}} recurrent neural network~\cite{hochreiter1997long} to obtain a continuous vector representation $h(x)$. This vector is then split into \emph{local} and \emph{global} components $h(x) = [h_1(x); h_2(x)]$. The local component,  $h_2(x)$, is reshaped into a kernel to perform a convolution operation on the state embedding $\phi(s)$ (similar to Chen et al. \shortcite{chen2015abc}):
\begin{dmath}
	z_1 = \psi_1(\phi(s); h_2(x))
\end{dmath}

Meanwhile, the three-element global component $h_1(x)$ is used to form the coefficients for a vertical and horizontal gradient along with a corresponding bias term.\footnote{Note that we are referring to gradient filters here, not the gradient calculated during backpropagation in deep learning.} The gradients, denoted $G_1$ and $G_2$ in Figure~\ref{fig:model}, are matrices of the same dimensionality as the state observation with values increasing down the rows and along the columns, respectively. The axis-aligned gradients are weighted by the elements of $h_1(x)$ and summed to give a final global gradient spanning the entire $2${\sc{D}} space, analogous to how steerable filters can be constructed for any orientation using a small set of basis filters \cite{freeman1991steerable}: 
\begin{dmath}
	z_2 = h_{1}(x)[1] \cdot G_1 + h_{1}(x)[2] \cdot G_2 + h_{1}(x)[3] \cdot J
\end{dmath}
in which $J$ is the all-ones matrix also of the same dimensionality as the observed map.

Finally, the local and global information maps are concatenated into a single tensor, which is then processed by a convolutional neural network (CNN) with parameters $\theta$ to approximate the generalized value function:
\begin{dmath}
	\hat{V}(s,x) = \psi_2([z_1; z_2]; \theta)
\end{dmath}
for every state $s$ in the map.


\begin{algorithm}[t]
\caption{Training Procedure}
\label{alg:training}
\small
\begin{algorithmic}[1]
\State Initialize experience memory $\mathcal{D}$ 
\State Initialize model parameters $\Theta$
\medmuskip=0mu
\thinmuskip=0mu
\thickmuskip=0mu
\For {$ epoch = 1,M $}
    \State Sample instruction $x \in X$ and associated environment $E$  
    \State Predict value map $\hat{V}(s,x;\Theta) \text{ for all } s \in E$
    \State Choose start state $s_0$ randomly
    \For {$ t = 1, N $ }
    	\State Select $a_t = \argmax\limits_{a} \sum\limits_{s} T(s | s_{t-1}, a)\hat{V}(s,x;\Theta)$
		\State Observe next state $s_t$ and reward $r_t$
    \EndFor
    \State Store trajectory  $(\bm{s}=s_0,s_1,\ldots, \bm{r}=r_0,r_1,\ldots)$ in $\mathcal{D}$
    
    \For {$j=1,J$} 
    	\State Sample random trajectory $(\bm{s}, \bm{r})$ from $\mathcal{D}$
		\State Perform gradient descent step on loss $\mathcal{L}(\theta)$
    \EndFor
\EndFor

\end{algorithmic}
\normalsize
\end{algorithm}

\paragraph{Reinforcement Learning}
Given our model's $\hat{V}(s,x)$ predictions, the resulting policy (Equation~\ref{eq:policy}) can be enacted, giving a continuous trajectory of states $\{s_t,s_{t+1},\ldots\}$ on a single map and their associated rewards $\{r_t,r_{t+1},\ldots\}$ at each timestep $t$. We stored entire trajectories (as opposed to state transition pairs) in a replay memory $\mathcal{D}$ as described in Mnih et al. \shortcite{mnih2015dqn}. The model is trained to produce an accurate value estimate by minimizing the following objective:

\begin{dmath}
	\mathcal{L}(\Theta) = \mathrm{E}_{s \sim \mathcal{D}}  \left[  \hat{V}(s,x;\Theta) - \left(R(s, x) + \\ \gamma \max_{a} \sum_{s'} T(s' | s, a) \hat{V}(s',x;\Theta^{-}) \right) \right] ^2 
\label{eq:loss}
\end{dmath}
where $s$ is a state sampled from $\mathcal{D}$, $\gamma$ is the discount factor, $\Theta$ is the set of parameters of the entire model, and $\Theta^{-}$ is the set of parameters of a target network copied periodically from our model. The complete training procedure is shown in Algorithm~\ref{alg:training}.



\section{Experimental Setup}
\label{sec:experiments}

\paragraph{Puddle world navigation data}

In order to study generalization across a wide variety of environmental conditions and linguistic inputs, we develop an extension of the \emph{puddle world} reinforcement learning benchmark \cite{sutton1995generalization,mankowitz2016ihomp}. States in a $10 \times 10$ grid are first filled with either grass or water cells, such that the grass forms one connected component. We then populate the grass region with six unique objects which appear only once per map (triangle, star, diamond, circle, heart, and spade) and four non-unique objects (rock, tree, horse, and house) which can appear any number of times on a given map. See Figure~\ref{fig:example} for an example visualization. 

\begin{table}[h]
\centering
\begin{tabular}{  c  c  c  } \toprule
Split & \textbf{Local} & \textbf{Global} \\ \midrule
Train & 1566 & 1071 \\
Test & 399 &  272 \\ \bottomrule
\end{tabular}
\caption{Overall statistics of our dataset.}
\label{table:data}
\end{table}

Goal positions are chosen uniformly at random from the set of grass cells, encouraging the use of spatial references to describe goal locations which do not themselves contain a unique object. We used the Mechanical Turk crowdsourcing platform~\cite{mturk} to collect natural language descriptions of these goals. Human annotators were asked to describe the positions of these goals using surrounding objects. At the end of each trial, we asked the same participants to provide goal locations given their own text instructions. This helped filter out a majority of instructions that were ambiguous or ill-specified. Table~\ref{table:data} provides some statistics on the data, and Figure~\ref{fig:annotations} shows example annotations. In total, we collected 3308 instructions, ranging from 2 to 43 words in length, describing over 200 maps. There are 361 unique words in the annotated instructions. We do not perform any preprocessing on the raw annotations.

\begin{figure}
  \begin{annotationbox}
    \small
      \begin{itemize}
        \item \emph{Reach the horse below the rock and to the left of the green diamond} 
        \item \emph{Move to the square two below and one left of the star} 
        \item \emph{Go to the cell above the bottommost horse}
      \end{itemize}
  \end{annotationbox}
  \caption{Example goal annotations collected with Mechanical Turk.}
  \label{fig:annotations}
\end{figure}

It is plausible that a model designed to handle only local references could not handle global ones (consider our own model without the global gradient maps). For clearer interpretation of results, we evaluate our model in two modes: trained and tested on local and global data separately, or as a combined dataset. While local instructions were obtained easily, the global instructions were collected by designing a task in which only nonunique objects were presented to the annotators.\footnote{The other objects were added back into the map after collecting the instruction.} This precluded simple instructions like \emph{``go left of the $\langle$object$\rangle$"} because there would always be more than one of each object type. Therefore, we obtained text with global properties (e.g. \emph{middle rock}, \emph{leftmost tree}) to sufficiently pinpoint an object. On average, we collected 31 unique local instructions and 10 unique global instructions per map.



\begin{figure}[t]
    \centering
    \includegraphics[width=0.8\linewidth]{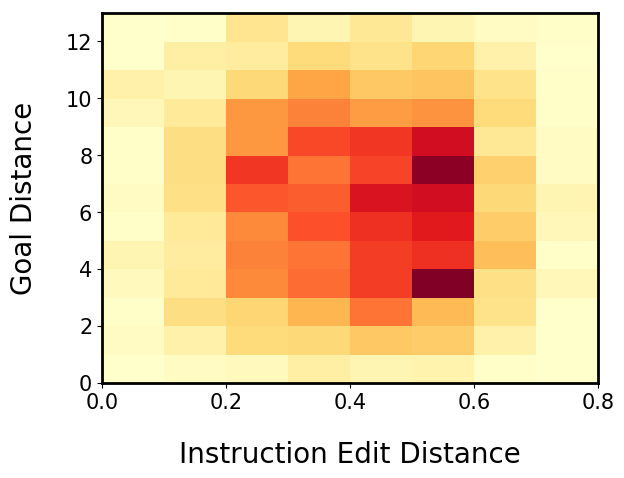}
    \caption{A heatmap showing the normalized instruction edit distance and goal Manhattan distance corresponding to the most similar instructions between the train and test set. For each instruction in the test set, we find the five most similar instructions in the training set. Even for those instructions which are similar, the goal locations they describe can be far apart.}
    \label{fig:similar_instructions}
\end{figure}

\begin{figure*}
    \centering
    \includegraphics[width=1.0\linewidth]{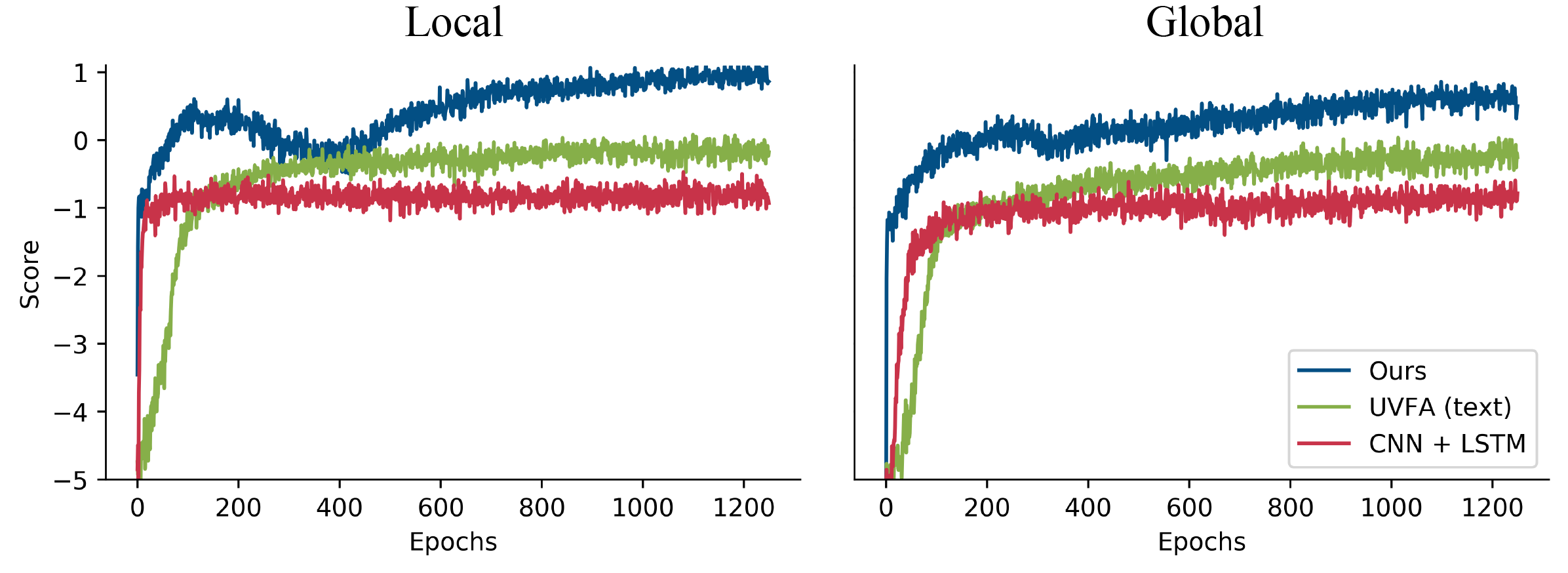}
    \caption{Reward achieved by our model and the two baselines on the training environments during reinforcement learning on both local and global instructions. Each epoch corresponds to simulation on 500 goals, with a goal simulation terminating either when the agent reaches the goal state or has taken 75 actions.}
    \label{fig:learning_curves}
\end{figure*}

\normalsize
To quantify the diversity of our dataset, we find the five nearest instructions in the training set for every instruction in the test set, as measured by edit distance (using the word as a unit) normalized by test instruction length. For each of these pairs, we also measure the Manhattan distance between their corresponding goal locations. Figure~\ref{fig:similar_instructions}, which visualizes this analysis, underscores the difficulty of this task; even when two instructions are highly similar, they might correspond to entirely different target locations. This is the case in the example in Figure~\ref{fig:example}, which has a distance of four between the references goals. 

\paragraph{Baselines}
\label{sec:baselines}
We compare our model to the following baselines:

\textbf{UVFA (text)} is a variant of the model described in \cite{schaul2015universal} adapted for our task. The original model made use of two MLPs to learn low dimensional embeddings of states and goals which were then combined via dot product to give value estimates. Goals were represented either as $(x,y)$ coordinates or as states themselves. As our goals are not observed directly but described in text, we replace the goal MLP with the same LSTM as in our model. The state MLP has an identical architecture to that of the UVFA: two hidden layers of dimension 128 and ReLU activations. For consistency with the UVFA, we represent states as binary vectors denoting the presence of each type of object at every position. 
 
\textbf{CNN + LSTM} is a variant of the model described in Misra et al. \shortcite{misra2017mapping}, who developed it for a language-grounded block manipulation task. It first convolves the map layout to a low-dimensional representation (as opposed to the MLP of the UVFA) and concatenates this to the LSTM's instruction embedding (as opposed to a dot product). These concatenated representations are then input to a two-layer MLP.

We also perform analysis to study the representational power of our model, introducing two more comparison models: 

 \textbf{UVFA (pos)} is the original UVFA model from \cite{schaul2015universal}, which we evaluate on our modified puddle worlds to determine the difficulty of environment generalization independently from instruction interpretation.
 
 \textbf{Our model (w/o gradient)} is an ablation of our model without the global gradient maps, which allows us to determine the gradients' role in representation-building.
 
In additional to our reinforcement learning experiments, we train these models in a supervised setting to isolate the effects of architecture choices from other concerns inherent to reinforcement learning algorithms. For this purpose, we constructed a dataset of ground-truth value maps for all human-annotated goals using value iteration. We use the models to predict value maps for the entire grid and minimize the mean squared error (MSE) compared to the ground truth values:

\begin{dmath}
\mathcal{L}'(\Theta) = \sum_{(s,x)} [\hat{V}(s,x; \Theta) - \bar{V}(s,x)]^2
\end{dmath}

\begin{table*}
  \renewcommand\arraystretch{1.3}
  \renewcommand\tabcolsep{0pt}
  \begin{tabular*}{1\linewidth}{@{\extracolsep{\fill}}lcccccc}
        \toprule
        & \multicolumn{2}{c}{Local} & \multicolumn{2}{c}{Global} & \multicolumn{2}{c}{Combined} \\
        \cmidrule(lr){2-3}\cmidrule(lr){4-5}\cmidrule(lr){6-7} 
        & Policy Quality & Distance
        & Policy Quality & Distance
        & Policy Quality & Distance \\
        \midrule
        UVFA (text) & 0.56 & 4.71 & 0.62 & 6.28 & 0.61 & 5.06 \\
        CNN + LSTM & 0.80 & 5.73 & 0.82 & 6.13 & 0.82 & 5.67 \\
        Our model & \textbf{0.87} & \textbf{2.18} & \textbf{0.90} & \textbf{3.35} & \textbf{0.89} & \textbf{3.04} \\
        \bottomrule
  \end{tabular*}
  \vspace{0.1cm}
  \caption{Performance of models trained via reinforcement learning on a held-out set of environments and instructions. Policy quality is the true expected normalized reward and distance denotes the Manhattan distance from goal location prediction to true goal position. We show results from training on the local and global instructions both separately and jointly.}
  \label{tbl:rl_eval}
\end{table*}

\begin{table*}[t]
  \renewcommand\arraystretch{1.3}
  \renewcommand\tabcolsep{0pt}
  \begin{tabular*}{1\linewidth}{@{\extracolsep{\fill}}lcccccc}
        \toprule
        & \multicolumn{3}{c}{Local} & \multicolumn{3}{c}{Global} \\
        \cmidrule(lr){2-4}\cmidrule(lr){5-7}
        & MSE & Policy Quality & Distance
        & MSE & Policy Quality & Distance \\
        \midrule
        UVFA (pos) & 1.30 & 0.48 & 4.96 & 1.04 & 0.52 & 5.39 \\
        UVFA (text) & 3.23 & 0.57 & 4.97 & 1.9 & 0.62 & 5.31 \\
        CNN + LSTM & 0.42 & 0.86 & 4.08 & 0.43 & 0.83 & 4.18 \\
        \midrule
        Our model (w/o gradient) & \textbf{0.25} & \textbf{0.94} & 2.39 & 0.61 & 0.87 & 5.15 \\
        Our model & \textbf{0.25} & \textbf{0.94} & \textbf{2.34} & \textbf{0.41} & \textbf{0.89} & \textbf{3.81}  \\
        \bottomrule
  \end{tabular*}
  \vspace{0.1cm}
  \caption{Performance on a test set of environments and instructions after supervised training. Lower is better for MSE and Manhattan distance; higher is better for policy quality. The gradient basis significantly improves the reconstruction error and goal localization of our model on global instructions, and expectedly does not affect its performance on local instructions.}
  \label{tbl:analysis} 
\end{table*}

\subsection{Implementation details}
Our model implementation uses an LSTM with a learnable 15-dimensional embedding layer, 30 hidden units, 8-dimensional embeddings $\phi(s)$, and a 3x3 kernel applied to the embeddings, giving a dimension of $72$ for $h_2(t)$. The final CNN has layers of $\{3,6,12,6,3,1\}$ channels, all with 3x3 kernels and padding of length 1 such that the output value map prediction is equal in size to the input observation.
For each map, a reward of $3$ is given for reaching the correct goal specified by human annotation and a reward of $-1$ is given for falling in a puddle cell. The only terminal state is when the agent is at the goal. Rewards are discounted by a factor of 0.95. We use Adam optimization~\cite{ba2014adam} for training all models.

\section{Results}
We present empirical results on two different datasets - our annotated puddle world and an existing block navigation task~\cite{bisk2016naacl}.

\subsection{Puddle world navigation}
\paragraph{Comparison with the state-of-the-art} 
We first investigate the ability of our model to learn solely from environment simulation. Figure~\ref{fig:learning_curves} shows the discounted reward achieved by our model as well as the two baselines for both instruction types. In both experiments, our model is the only one of the three to achieve an average nonnegative reward after convergence (0.88 for local instructions and 0.49 for global instructions), signifying that the baselines do not fully learn how to navigate through these environments. 

Following Schaul et al. \shortcite{schaul2015universal}, we also evaluated our model using the metric of \emph{policy quality}. This is defined as the expected discounted reward achieved by following a softmax policy of the value predictions. 
Policy quality is normalized such that an optimal policy has a score of 1 and a uniform random policy has a score of 0.
Intuitively, policy quality is the true normalized expectation of score over all maps in the dataset, instructions per map, and start states per map-instruction pair.  Our model outperforms both baselines on this metric as well on the test maps (Table~\ref{tbl:rl_eval}). We also note that the performance of the baselines flip with respect to each other as compared to their performance on the training maps (Figure \ref{fig:learning_curves}). While the UVFA variant learned a better policy on the train set, it did not generalize to new environments as well as the CNN + LSTM. 

Finally, given the nature of our environments, we can use the predicted value maps to infer a goal location by taking the position of the maximum value. We use the Manhattan distance from this predicted position to the actual goal location as a third metric. The accuracy of our model's goal predictions is more than twice that of the baselines on local references and roughly $45\%$ better on global references. 


\begin{figure*}[!h]
    \centering
    \includegraphics[width=\linewidth]{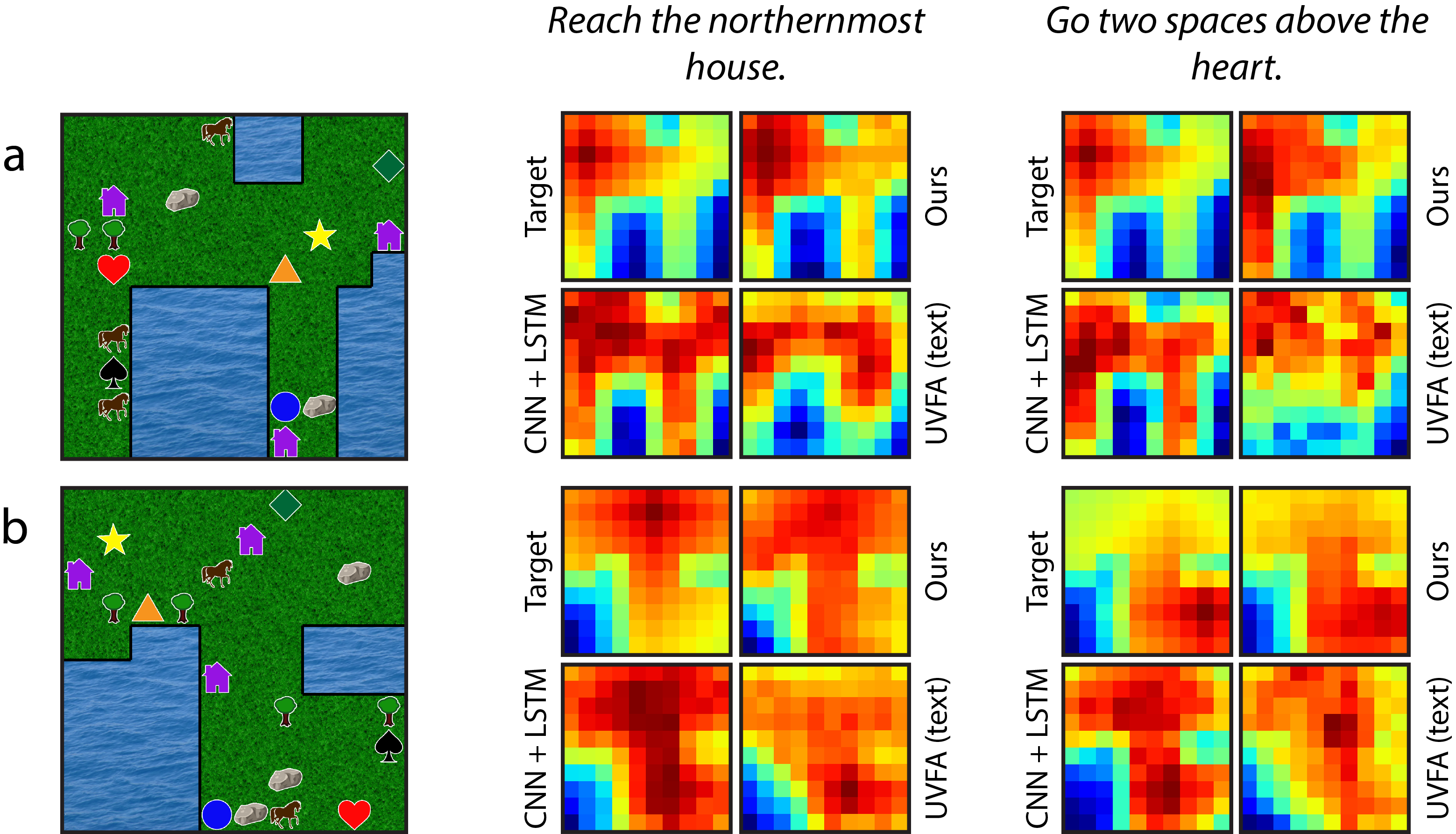}
    \caption{Value map predictions for two environments paired with two instructions each. Despite the difference in instructions, with one being global and the other local in nature and sharing no objects in their descriptions, they refer to the same goal location in the environment in (a). However, in (b), the descriptions correspond to different locations on the map. The vertical axis considers variance in goal location for the same instruction, depending on the map configuration.}
    \label{fig:variations}
\end{figure*}

\begin{figure}[!h]
    \centering
    \includegraphics[width=0.9\linewidth]{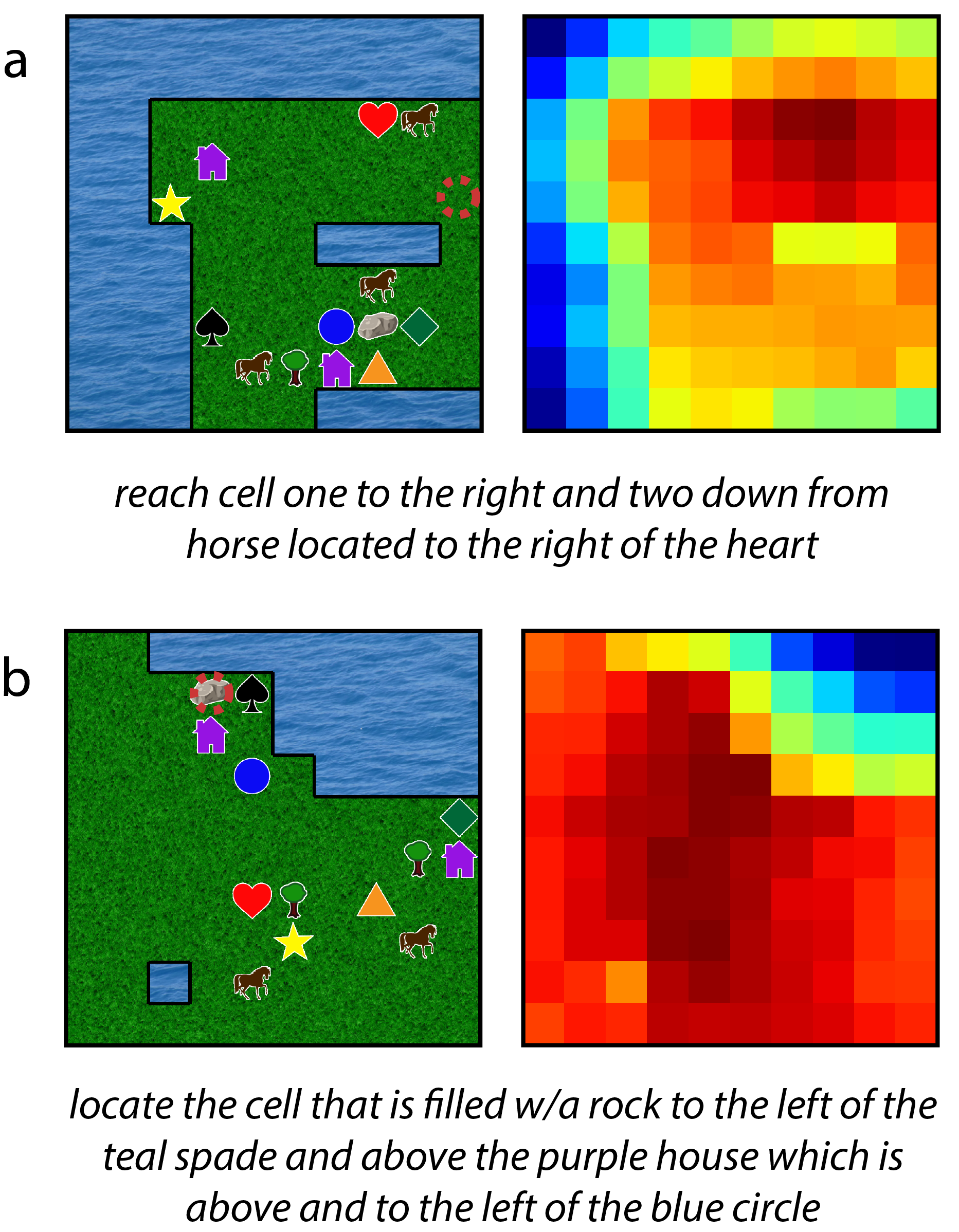}
    \caption{Examples of failure cases for our model. Multiple levels of indirection in (a) and a long instruction filled with redundant information in (b) make the instruction difficult to interpret. Intended goal locations are outlined in red for clarity. }
    \label{fig:qualitative}
\end{figure}

\begin{figure}
    \centering
    \includegraphics[width=0.8\linewidth]{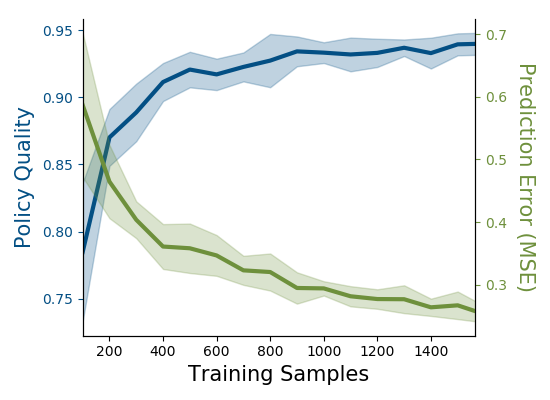}
    \caption{Effect of training set size on held-out predictions. The curves show the mean of ten training runs and the shaded regions show standard deviation. Our model's policy quality is greater than 0.90 with as few as 400 training goal annotations.}
    \label{fig:training_size}
\end{figure}

\paragraph{Analysis of learned representations}
For the representation analysis in a supervised setting, we compared the predicted value maps of all models against the unseen test split of maps. 
Table~\ref{tbl:analysis} shows the results of this study. As expected, our model without the global gradient performs no differently from the full model on local references, but has higher MSE and average distances to true goal than the full model on global references. We also note that UVFA (pos) performs much worse than both CNN+LSTM and our model, showing the difficulty of environment generalization even when the goals are observed directly. (The original UVFA paper~\cite{schaul2015universal} demonstrated effective generalization over goal states within a \emph{single} environment.)

Surprisingly, our model trained via reinforcement learning has more precise goal location predictions (as measured via Manhattan distance) than when trained on true state values in a supervised manner. However, the MSE of the value predictions are much higher in the RL setting (\emph{e.g.}, 0.80 vs 0.25 for supervised on local instructions). This shows that despite the comparative stability of the supervised setting, minimization of value prediction error does not necessarily lend itself to the best policy or goal localization. Conversely, having a higher MSE does not always imply a worse policy, as seen also in the performance of the two UVFA variants in Table~\ref{tbl:analysis}. 

\paragraph{Generalization} 
One of the criteria laid out for our model was its ability to construct language representations and produce accurate value maps, independent of layouts and linguistic variation. Figure~\ref{fig:variations} provides examples of two layouts, each with two different instructions. In the first map (top), we have both instructions referring to the same location. Our model is able to mimic the optimal value map accurately, while the other baselines are not as precise, either producing a large field of possible goal locations (\emph{CNN+LSTM}) or completely missing the goal (\emph{UVFA-text}).

On the vertical axis, we observe generalization across different maps with the same instructions. Our model is able to precisely identify the goals in each scenario in spite of significant variation in their locations. This proves harder for the other representations.

Although our model is compositional in the sense that it transfers knowledge of spatial references between different environments, some types of instructions do prove challenging. We identify two of the poorest predictions in Figure~\ref{fig:qualitative}. We see that multiple levels of indirection (as in \ref{fig:qualitative}a, which references a location relative to an object relative to another object) or unnecessarily long instructions (as in \ref{fig:qualitative}b, which uniquely identifies a position by the eighth token but then proceeds with redundant information) are still a challenge.

\paragraph{Learning curve} 
Due to the manual effort that comes with constructing a dataset of human annotations, it is also important to consider the sample-efficiency of a model. Figure~\ref{fig:training_size} shows the quality policy and prediction error on local instructions as a function of training set size. We observe that our model reaches 0.90 policy quality with only 400 samples, demonstrating efficient generalization capability.


\begin{figure*}[h!]
    \centering
    \includegraphics[width=\linewidth]{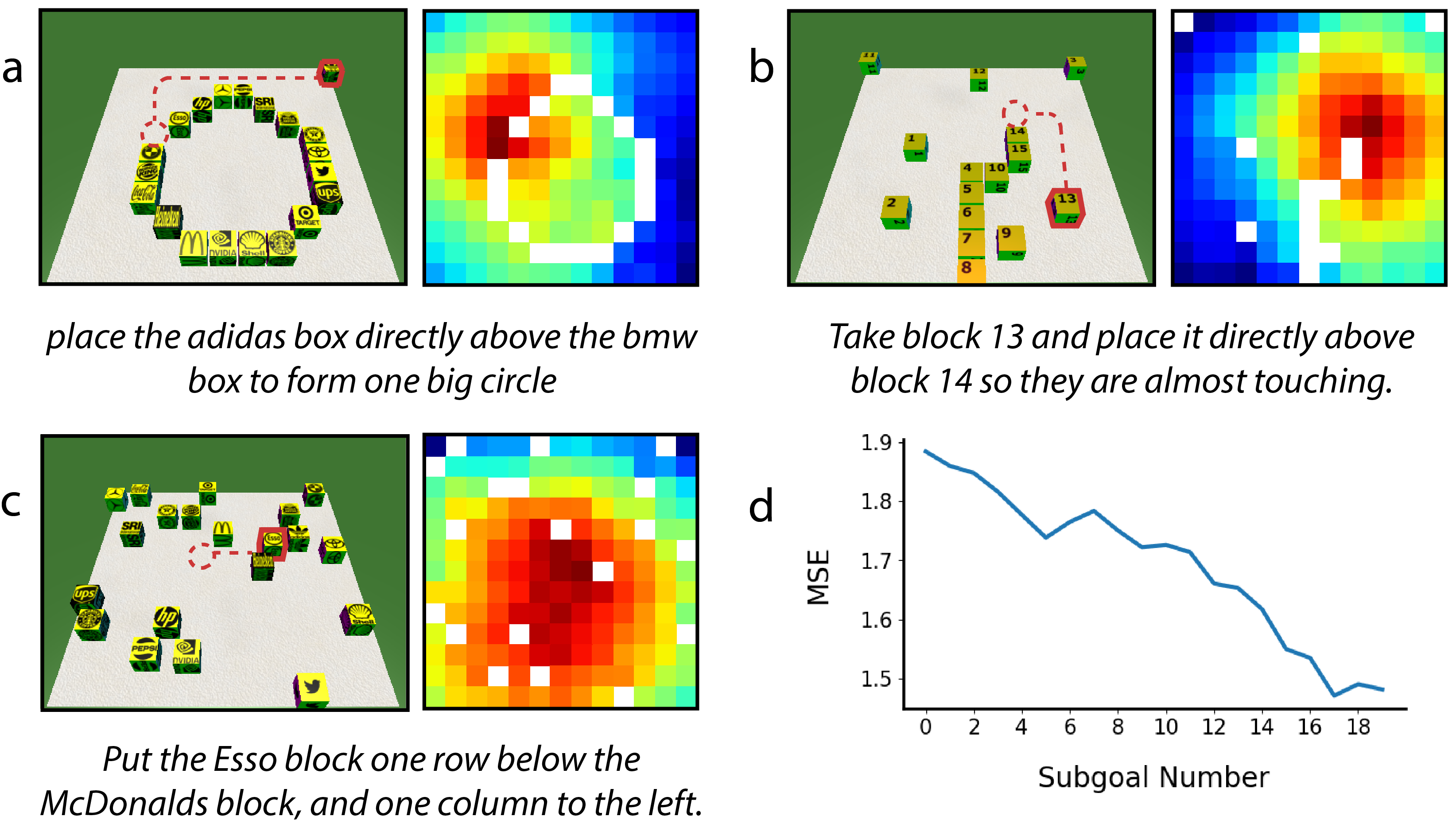}
    \caption{(a-c) Visualizations of tasks from the ISI Language Grounding dataset \protect\cite{bisk2016naacl} and our model's value map predictions. The agentive block and goal location are outlined in red for visibility. (d) The MSE of the value map prediction as a function of a subgoal's ordering in an overall task. The model performs better on subgoals later in a task despite the subgoals being treated completely independently during both training and testing.}
    \label{fig:bisk}
\end{figure*}

\begin{table}[h!]
  \renewcommand\arraystretch{1.3}
  \renewcommand\tabcolsep{0pt}
  \begin{tabular*}{1\linewidth}{@{\extracolsep{\fill}}lcc}
        \toprule
        & \multicolumn{2}{c}{Language Grounding Dataset}  \\
        \cmidrule(lr){2-3}
        & Policy Quality & Distance \\
        \midrule
        UVFA (text) & 0.15 & 4.61  \\
        CNN + LSTM & 0.17 & 4.11 \\
        Our model & \textbf{0.74} & \textbf{3.94} \\
        \bottomrule
  \end{tabular*}
  \vspace{0.1cm}
  \caption{The performance of our model and two baselines on the ISI Language Grounding dataset \protect\cite{bisk2016naacl}. Our model once again outperforms the baselines, although all models have a lower policy quality on this dataset than on our own.}
  \label{tbl:bisk}
\end{table}

\subsection{ISI Grounding Dataset} 
We also evaluate our model on the ISI Language Grounding dataset \cite{bisk2016naacl}, which contains human-annotated instructions describing how to arrange blocks identified by numbers and logos. Although it does not contain variable environment maps as in our dataset, it has a larger action space and vocabulary. The caveat is that the task as posed in the original dataset is not compatible with our model. For a policy to be derived from a value map with the same dimension as the state observation, it is implicitly assumed that there is a single controllable agent, whereas the ISI set allows multiple blocks to be moved. We therefore modify the ISI setup using an oracle to determine which block is given agency during each step. This allows us to retain the linguistic variability of the dataset while overcoming the mismatch in task setup. The states are discretized to a $13 \times 13$ map and the instructions are lemmatized.

Performance on the modified ISI dataset is reported in Table~\ref{tbl:bisk} and representative visualizations are shown in Figure~\ref{fig:bisk}. Our model outperforms both baselines by a greater margin in policy quality than on our own dataset.






Misra et al. \shortcite{misra2017mapping} also use this dataset and report results in part by determining the minimum distance between an agent and a goal during an evaluation lasting $N$ steps. This evaluation metric is therefore dependent on this timeout parameter $N$. Because we discretized the state space so as to be able to represent it as a grid of embeddings, the notion of a single step has been changed and direct comparison limited to $N$ steps is ill-defined.\footnote{When a model is available and the states are not overwhelmingly high-dimensional, policy quality is a useful metric that is independent of this type of parameter. As such, it is our default metric here. However, estimating policy quality for environments substantially larger than those investigated here is a challenge in itself.} Hence, due to modifications in the task setup, we cannot compare directly to the results in Misra et al. \shortcite{misra2017mapping}.





\paragraph{Understanding grounding evaluation}
An interesting finding in our analysis was that the difficulty of the language interpretation task is a function of the stage in task execution (Figure~\ref{fig:bisk}(d)).  In the ISI Language Grounding set \cite{bisk2016naacl}, each individual instruction (describing where to move a particular block) is a subgoal in a larger task (such as constructing a circle with all of the blocks). The value maps predicted for subgoals occurring later in a task are more accurate than those occurring early in the task. 
It is likely that the language plays a less crucial role in specifying the subgoal position in the final steps of a task.
As shown in Figure~\ref{fig:bisk}(a), it may be possible to narrow down candidate subgoal positions just by looking at a nearly-constructed high-level shape. In contrast, this would not be possible early in a task because most of the blocks will be randomly positioned. This finding is consistent with a result from Branavan et al. \shortcite{branavan2011learning}, who reported that strategy game manuals were useful early in the game but became less essential further into play. It appears to be part of a larger trend that the marginal benefit of language in such grounding tasks can vary predictably between individual instructions. 












\section{Conclusions}
\label{sec:conclusions}

We have described a novel approach for grounded spatial reasoning. Combining the language representation in a spatially localized manner allows for increased precision of goal identification and improved performance on unseen environment configurations. Alongside our models, we present \emph{Puddle World Navigation}, a new grounding dataset for testing the generalization capacity of instruction-following algorithms in varied environments. 

\section*{Acknowledgement}
We thank the members of the MIT NLP group for helpful discussions and feedback. We gratefully acknowledge support from the MIT Lincoln Laboratory and the MIT SuperUROP program.
\bibliography{references}

\begin{thebibliography}{}

\bibitem[\protect\citename{Andreas and Klein}2015]{Andreas15Instructions}
Jacob Andreas and Dan Klein.
\newblock 2015.
\newblock Alignment-based compositional semantics for instruction following.
\newblock In {\em EMNLP}.

\bibitem[\protect\citename{Artzi and Zettlemoyer}2013]{artzi2013weakly}
Yoav Artzi and Luke Zettlemoyer.
\newblock 2013.
\newblock Weakly supervised learning of semantic parsers for mapping
  instructions to actions.
\newblock {\em TACL}, 1(1):49--62.

\bibitem[\protect\citename{Bellman}1957]{bellman1957}
Richard Bellman.
\newblock 1957.
\newblock {\em Dynamic Programming}.
\newblock Princeton University Press, Princeton, NJ, USA, 1 edition.

\bibitem[\protect\citename{Bisk \bgroup et al.\egroup }2016]{bisk2016naacl}
Yonatan Bisk, Deniz Yuret, and Daniel Marcu.
\newblock 2016.
\newblock Natural language communication with robots.
\newblock In {\em {NAACL} {HLT}}, pages 751--761.

\bibitem[\protect\citename{Branavan \bgroup et al.\egroup
  }2011]{branavan2011learning}
SRK Branavan, David Silver, and Regina Barzilay.
\newblock 2011.
\newblock Learning to win by reading manuals in a monte-carlo framework.
\newblock In {\em ACL HLT}, pages 268--277.

\bibitem[\protect\citename{Buhrmester \bgroup et al.\egroup }2011]{mturk}
Michael Buhrmester, Tracy Kwang, and Samuel~D. Gosling.
\newblock 2011.
\newblock Amazon's mechanical turk.
\newblock {\em Perspectives on Psychological Science}, 6(1):3--5.
\newblock PMID: 26162106.

\bibitem[\protect\citename{Byrne and Johnson-Laird}1989]{byrne1989spatial}
Ruth~MJ Byrne and Philip~N Johnson-Laird.
\newblock 1989.
\newblock Spatial reasoning.
\newblock {\em Journal of memory and language}, 28(5):564--575.

\bibitem[\protect\citename{Chen and Mooney}2011]{chen2011learning}
David~L Chen and Raymond~J Mooney.
\newblock 2011.
\newblock Learning to interpret natural language navigation instructions from
  observations.
\newblock {\em San Francisco, CA}, pages 859--865.

\bibitem[\protect\citename{Chen \bgroup et al.\egroup }2015]{chen2015abc}
Kan Chen, Jiang Wang, Liang-Chieh Chen, Haoyuan Gao, Wei Xu, and Ram Nevatia.
\newblock 2015.
\newblock Abc-cnn: An attention based convolutional neural network for visual
  question answering.
\newblock {\em arXiv preprint arXiv:1511.05960}.

\bibitem[\protect\citename{Freeman and Adelson}1991]{freeman1991steerable}
William~T. Freeman and Edward~H. Adelson.
\newblock 1991.
\newblock The design and use of steerable filters.
\newblock {\em IEEE TPAMI}, 13(9):891--906, September.

\bibitem[\protect\citename{Hemachandra \bgroup et al.\egroup
  }2014]{hemachandra2014learning}
Sachithra Hemachandra, Matthew~R Walter, Stefanie Tellex, and Seth Teller.
\newblock 2014.
\newblock Learning spatial-semantic representations from natural language
  descriptions and scene classifications.
\newblock In {\em ICRA}, pages 2623--2630. IEEE.

\bibitem[\protect\citename{Hochreiter and Schmidhuber}1997]{hochreiter1997long}
Sepp Hochreiter and J{\"u}rgen Schmidhuber.
\newblock 1997.
\newblock Long short-term memory.
\newblock {\em Neural computation}, 9(8):1735--1780.

\bibitem[\protect\citename{Kelleher \bgroup et al.\egroup
  }2006]{kelleher2006proximity}
John~D Kelleher, Geert-Jan~M Kruijff, and Fintan~J Costello.
\newblock 2006.
\newblock Proximity in context: an empirically grounded computational model of
  proximity for processing topological spatial expressions.
\newblock In {\em ACL}, pages 745--752.

\bibitem[\protect\citename{Kim and Mooney}2013]{kim2013adapting}
Joohyun Kim and Raymond~J Mooney.
\newblock 2013.
\newblock Adapting discriminative reranking to grounded language learning.
\newblock In {\em ACL (1)}, pages 218--227.

\bibitem[\protect\citename{Kingma and Ba}2015]{ba2014adam}
Diederik~P. Kingma and Jimmy Ba.
\newblock 2015.
\newblock Adam: A method for stochastic optimization.
\newblock In {\em ICLR}.

\bibitem[\protect\citename{Kollar \bgroup et al.\egroup
  }2010]{kollar2010toward}
Thomas Kollar, Stefanie Tellex, Deb Roy, and Nicholas Roy.
\newblock 2010.
\newblock Toward understanding natural language directions.
\newblock In {\em Human-Robot Interaction}, pages 259--266. IEEE.

\bibitem[\protect\citename{Levinson}2003]{levinson2003space}
Stephen~C Levinson.
\newblock 2003.
\newblock {\em Space in language and cognition: Explorations in cognitive
  diversity}, volume~5.
\newblock Cambridge University Press.

\bibitem[\protect\citename{Levit and Roy}2007]{levit2007interpretation}
Michael Levit and Deb Roy.
\newblock 2007.
\newblock Interpretation of spatial language in a map navigation task.
\newblock {\em IEEE Transactions on Systems, Man, and Cybernetics, Part B
  (Cybernetics)}, 37(3):667--679.

\bibitem[\protect\citename{Li and Gleitman}2002]{li2002turning}
Peggy Li and Lila Gleitman.
\newblock 2002.
\newblock Turning the tables: Language and spatial reasoning.
\newblock {\em Cognition}, 83(3):265--294.

\bibitem[\protect\citename{MacMahon \bgroup et al.\egroup
  }2006]{macmahon2006walk}
Matt MacMahon, Brian Stankiewicz, and Benjamin Kuipers.
\newblock 2006.
\newblock Walk the talk: Connecting language, knowledge, and action in route
  instructions.
\newblock {\em AAAI}, 2(6):4.

\bibitem[\protect\citename{Mankowitz \bgroup et al.\egroup
  }2016]{mankowitz2016ihomp}
Daniel~J. Mankowitz, Timothy~Arthur Mann, and Shie Mannor.
\newblock 2016.
\newblock Iterative hierarchical optimization for misspecified problems
  {(IHOMP)}.
\newblock {\em CoRR}, abs/1602.03348.

\bibitem[\protect\citename{Misra \bgroup et al.\egroup }2017]{misra2017mapping}
Dipendra~K Misra, John Langford, and Yoav Artzi.
\newblock 2017.
\newblock Mapping instructions and visual observations to actions with
  reinforcement learning.
\newblock {\em EMNLP}.

\bibitem[\protect\citename{Mnih \bgroup et al.\egroup }2015]{mnih2015dqn}
Volodymyr Mnih, Koray Kavukcuoglu, David Silver, Andrei~A. Rusu, Joel Veness,
  Marc~G. Bellemare, Alex Graves, Martin Riedmiller, Andreas~K. Fidjeland,
  Georg Ostrovski, Stig Petersen, Charles Beattie, Amir Sadik, Ioannis
  Antonoglou, Helen King, Dharshan Kumaran, Daan Wierstra, Shane Legg, and
  Demis Hassabis.
\newblock 2015.
\newblock Human-level control through deep reinforcement learning.
\newblock {\em Nature}, 518(7540):529--533, 02.

\bibitem[\protect\citename{Moratz and Tenbrink}2006]{moratz2006spatial}
Reinhard Moratz and Thora Tenbrink.
\newblock 2006.
\newblock Spatial reference in linguistic human-robot interaction: Iterative,
  empirically supported development of a model of projective relations.
\newblock {\em Spatial cognition and computation}, 6(1):63--107.

\bibitem[\protect\citename{Schaul \bgroup et al.\egroup
  }2015]{schaul2015universal}
Tom Schaul, Daniel Horgan, Karol Gregor, and David Silver.
\newblock 2015.
\newblock Universal value function approximators.
\newblock In {\em ICML}, pages 1312--1320.

\bibitem[\protect\citename{Skubic \bgroup et al.\egroup
  }2004]{skubic2004spatial}
M.~Skubic, D.~Perzanowski, S.~Blisard, A.~Schultz, W.~Adams, M.~Bugajska, and
  D.~Brock.
\newblock 2004.
\newblock Spatial language for human-robot dialogs.
\newblock {\em IEEE Transactions on Systems, Man, and Cybernetics, Part C
  (Applications and Reviews)}, 34(2):154--167, May.

\bibitem[\protect\citename{Sutton and Barto}1998]{sutton1998introduction}
Richard~S Sutton and Andrew~G Barto.
\newblock 1998.
\newblock {\em Introduction to reinforcement learning}.
\newblock MIT Press.

\bibitem[\protect\citename{Sutton}1996]{sutton1995generalization}
Richard~S. Sutton.
\newblock 1996.
\newblock Generalization in reinforcement learning: Successful examples using
  sparse coarse coding.
\newblock In {\em NIPS}.

\bibitem[\protect\citename{Tellex \bgroup et al.\egroup
  }2011]{tellex2011understanding}
Stefanie Tellex, Thomas Kollar, Steven Dickerson, Matthew~R Walter, Ashis~Gopal
  Banerjee, Seth~J Teller, and Nicholas Roy.
\newblock 2011.
\newblock Understanding natural language commands for robotic navigation and
  mobile manipulation.
\newblock In {\em AAAI}.

\bibitem[\protect\citename{Tenbrink}2007]{tenbrink2007space}
Thora Tenbrink.
\newblock 2007.
\newblock {\em Space, time, and the use of language: An investigation of
  relationships}, volume~36.
\newblock Walter de Gruyter.

\bibitem[\protect\citename{Viethen and Dale}2008]{viethen2008use}
Jette Viethen and Robert Dale.
\newblock 2008.
\newblock The use of spatial relations in referring expression generation.
\newblock In {\em Proceedings of the Fifth International Natural Language
  Generation Conference}, pages 59--67. Association for Computational
  Linguistics.

\bibitem[\protect\citename{Vogel and Jurafsky}2010]{vogel2010learning}
Adam Vogel and Dan Jurafsky.
\newblock 2010.
\newblock Learning to follow navigational directions.
\newblock In {\em ACL}, pages 806--814.

\bibitem[\protect\citename{Walter \bgroup et al.\egroup
  }2013]{walter2013learning}
Matthew~R Walter, Sachithra Hemachandra, Bianca Homberg, Stefanie Tellex, and
  Seth Teller.
\newblock 2013.
\newblock Learning semantic maps from natural language descriptions.
\newblock Robotics: Science and Systems.

\end{thebibliography}
\bibliographystyle{acl2012}

\end{document}